\documentclass[11pt]{article}

\usepackage[final]{acl}
\usepackage{times}
\usepackage{latexsym}
\usepackage{soul}
\usepackage[T1]{fontenc}
\usepackage[utf8]{inputenc}
\usepackage{microtype}
\usepackage{inconsolata}
\usepackage{graphicx}
\usepackage{xcolor}
\usepackage{amsmath}
\usepackage{multirow}
\usepackage{booktabs}
\usepackage{tcolorbox}
\usepackage{pifont}
\usepackage{xspace}

\newtheorem{definition}{Definition}

\newcommand{\cmark}{\ding{51}}
\newcommand{\xmark}{\ding{55}}
\def\dataset{DNE-ElecDeb\xspace}

\newtcolorbox{coloredbox}[3][]{
    colback=#3!5!white,
    colframe=#3!75!black,
    fonttitle=\fontsize{10pt}{10pt}\selectfont,
    fontupper=\fontsize{9pt}{10pt}\selectfont,
    fontlower=\scshape,
    title={#2},
    #1
}

\title{Who Argues What? Joint Argument--Entity Detection and Classification in Political Debates}

\author{Lucio La Cava, Stefano Francesco Monea, Sergio Greco \\
  DIMES Dept., University of Calabria \\
  v. P. Bucci 44Z, 87036 Rende, CS, Italy\\ 
  \texttt{\{lucio.lacava, sf.monea, greco\}@dimes.unical.it} \\ 
}

\begin{document}
\maketitle
\begin{abstract}
Political debates are often analyzed through Argument Mining (AM) to investigate the key arguments that drive them. 
However, political arguments are rarely interpretable from argumentative spans alone, as claims and premises generally depend on the entities (e.g., people, events, locations, parties) they mention. Existing AM resources and methods typically annotate argumentative spans and roles, but do not provide a paired debate-entity layer for asking which Debate Named Entities (DNE), e.g., actors and events, are invoked within debates.
In this work, we address these data and methodological gaps by (i) introducing \dataset, an entity-enriched version of the USElecDeb dataset that adds DNEs in both argumentative and non-argumentative spans and defines Debate Named Entity Recognition (DNER) as the task of detecting DNEs, and (ii) proposing Joint Argument and Entity Tagging (JAET), a generative framework that fine-tunes decoder-only LLMs to insert inline argument and entity tags into debate turns while preserving the original transcript.
Under BIO-tagging evaluation, JAET improves relative $F_1$ on the joint AM+DNER task by +27.3\%, resp. +41.9\%, under the untyped, resp. typed setting over the strongest sequential AM-DNER pipelines,  demonstrating that such gains cannot be recovered by composing two independent modules.
Notably, similar margins replicate on Persuasive Essays (+26.6\%, resp. +52.7\%), showing effective generalization to domains orthogonal to political debates.
By unifying argumentative and entity-level representations within a single view, our contributions pave the way for richer political debates understanding.
\end{abstract}

\section{Introduction}
Argument Mining (AM) aims to identify and interpret argumentative structures in natural language, unstructured texts~\cite{lawrence-reed-2019-argument}.
By automatically extracting argument components such as premises and claims, and by identifying the relations between them, AM enables the structured representation and analysis of arguments using formal argumentation formalisms~\cite{Dung95,aba,aspic}.

These representations support a wide range of downstream applications, including legal analysis~\cite{Palau2009argumentation}, scientific debate~\cite{sukpanichnant2024peerarg}, online discourse analysis~\cite{habernal-gurevych-2017-argumentation}, and political communication~\cite{goffredo-etal-2023-argument}. 

Political debates represent one of the most challenging settings for AM~\cite{Cabrio2018FiveYO}: they are not a mere collection of isolated and independent arguments, but interleaved turn-based conversational interactions between speakers who repeatedly refer to opponents, parties, institutions, and organizations.
Notably, these entities not only occur inside non-argumentative spans, e.g., mentioned by a moderator during the opening, but also permeate speakers' arguments, e.g., when a claim targets a specific actor or event.  

Traditional AM annotation frameworks for political debates capture the argumentative spans and roles, but they typically do not explicitly model the entities that contribute to making an argument (politically) meaningful. This is a key limitation: knowing that a textual span is a claim is informative, but knowing which entities are involved in a claim makes the annotation more useful for debate analysis. Indeed, entities become explicit semantic anchors for interpreting political debate arguments. 

This point becomes even more relevant considering that debates would typically benefit from streaming-compatible analysis. Without an explicit entity layer, traditional AM annotation frameworks can detect that speakers are making claims or premises, but they cannot reliably identify whether these involve the same set of entities. Consequently, they miss valuable parts of the debate structure needed to compare argumentative positions. 

Generic Named Entity Recognition (NER) alone cannot address this gap, as it recovers entities while completely missing the argumentative structure in which they appear. 

Similarly, running AM and NER as subsequent yet independent modules remains unsatisfactory: arguments and entities are intertwined, and separate processing can lose mutual dependencies or even produce incompatible boundaries. This motivates a unified formulation in which argument and entity annotations are learned and predicted together rather than merged.

Large Language Models (LLMs) are well-suited to this setting, as they can be fine-tuned to preserve the original discourse flow while inserting entity and argument markers in an autoregressive way.

\paragraph{Contributions}
To fill this gap, in this work, we make the following contributions:
\begin{itemize}
    \item We curate and release \dataset, an entity-enriched version of the \textit{USElecDeb60To20} dataset~\cite{haddadan-etal-2019-yes, goffredo-etal-2023-argument}, extending all 44 manually annotated debates with a paired debate-relevant entity layer covering both argumentative and non-argumentative spans;
    \item We propose \textit{Joint Argument and Entity Tagging} (JAET), a single-pass approach aimed at jointly predicting argument component boundaries, argument component labels, entity boundaries, and entity types. 
    JAET operates at a turn level, enabling streaming-compatible mining of debate structures as soon as speakers' turns arrive;
    \item We evaluate a representative set of small, open-weight LLMs, showing that JAET improves relative $F_1$ by +27.3\%, resp. +41.9\%, under the untyped, resp. typed, setting for joint AM+DNER tagging over the strongest sequential pipeline built on the same backbone and up to +24\%, resp. +36.4\%, for AM-only tagging, resp. DNER, over the strongest non-JAET baseline available for each task.
\end{itemize}

\section{Related Work}
\label{sec:related}
Argument Mining addresses various subtasks, including argument component segmentation (ACS) and classification (ACC), argument relation identification (ARI), and argument relation classification (ARC)~\cite{Cabrio2018FiveYO}. 

Early AM approaches leveraged feature-rich supervised methods such as maximum entropy classifiers~\cite{Palau11argmining}, logistic regression~\cite{levy-etal-2014-context}, Support Vector Machines~\cite{stab-gurevych-2014-identifying,niculae-etal-2017-argument}, and optimization techniques~\cite{stab2017persuasiveessay}. 
Neural architectures such as RNNs~\cite{eger-etal-2017-neural,niculae-etal-2017-argument}, LSTMs~\cite{potash-etal-2017-heres}, and Transformer-based models~\cite{mayer2020ecai,kashefi-etal-2023-argument,ding-etal-2022-dont,RooseBERT} improved AM capabilities by capturing richer contextual representations from argumentative texts.

Political debates have been widely recognized as a natural setting for AM~\cite{Cabrio2018FiveYO}. On the one hand, researchers introduced corpora of U.S. presidential campaign debates annotated with argument components and corresponding labels~\cite{haddadan-etal-2019-yes}, as well as fallacy annotations~\cite{goffredo-etal-2023-argument,goffredo-etal-2025-disputool} and social reactions~\cite{Visser2020datasety}.
On the other hand, a body of works assessed and applied AM techniques to political debates. Among these, \citet{LippiT16} detected claims in political debates, \citet{cano-basave-he-2016-study} investigated the impact of argumentative style in influencing an audience in supporting candidates, and \citet{MeniniCTV18} predicted relations between arguments in political speeches. However, all these resources and works focus on argument components and their relations, overlooking the entity layer.

\begin{table*}[t!]
\centering
\scriptsize
\setlength{\tabcolsep}{2.4pt}
\resizebox{\linewidth}{!}{
\begin{tabular}{llccccccc}
\toprule
 & \multirow{2}{*}{\textbf{Work}} & \multicolumn{2}{c}{\textbf{Data}} & \multicolumn{5}{c}{\textbf{Tasks}} \\
\cmidrule(lr){3-4}\cmidrule(lr){5-9}
 & & \textbf{Debates} & \textbf{Release} & \textbf{ACS} & \textbf{ACC} & \textbf{ACS+ACC} & \textbf{NER} & \textbf{ACS+ACC+NER} \\
\midrule
\multirow{7}{*}{\rotatebox[origin=c]{90}{\textbf{Argum. Mining}}}
& \citet{haddadan-etal-2019-yes} & \cmark & \cmark & \cmark & \cmark & \xmark & \xmark & \xmark \\
& \citet{LippiT16} & \cmark & \xmark & \cmark & \xmark & \xmark & \xmark & \xmark \\
& \citet{liu-etal-2023-argument} & \xmark & \xmark & \xmark & \cmark & \xmark & \xmark & \xmark \\
& \citet{cabessa-etal-2025-argument} & \xmark & \xmark & \xmark & \cmark & \xmark & \xmark & \xmark \\
& \citet{caputo-etal-2026-argument} & \xmark & \xmark & \cmark & \cmark & \xmark & \xmark & \xmark \\
& \citet{favero2025leveragingsmallllmsargument} & \xmark & \xmark & \cmark & \cmark & \xmark & \xmark & \xmark \\
& \citet{elguendouze2026compactpromptinginstructiontunedllms} & \cmark & \xmark & \cmark & \cmark & \cmark & \xmark & \xmark \\
\midrule
\multirow{2}{*}{\rotatebox[origin=c]{90}{\textbf{NER}}}
& GPT-NER~\cite{wang-etal-2025-gpt} & \xmark & \xmark & \xmark & \xmark & \xmark & \cmark & \xmark \\
& Universal-NER~\cite{Zhou00CP24} & \xmark & \xmark & \xmark & \xmark & \xmark & \cmark & \xmark \\
\midrule
\multirow{2}{*}{\rotatebox[origin=c]{90}{\textbf{Both}}}
& RooseBERT~\cite{RooseBERT} & \cmark & \xmark & \cmark$^*$ &\cmark$^*$ & \cmark$^*$ &\cmark$^*$ & \cmark$^*$ \\
& \textbf{\dataset + JAET} & \cmark & \cmark & \cmark & \cmark & \cmark & \cmark & \cmark \\
\bottomrule
\end{tabular}
}
\caption{Comparison with related AM/NER literature. \textit{Debates} = whether the work uses debate data, \textit{Release} = whether the work introduces or extends an annotated dataset relevant to its task, \textit{ACS} = Argument Component Segmentation, \textit{ACC} = Argument Component Classification, \textit{ACS+ACC} = joint ACS and ACC, \textit{NER} = (Generic) Named Entity Recognition, \textit{ACS+ACC+NER} = joint prediction of argument and entity boundary and types.
Asterisks indicate RooseBERT requiring task-specific fine-tuning before usage.
}
\label{tab:related-comparison}
\vspace{-2mm}
\end{table*}

The advent of LLMs brought new capabilities to AM by reformulating all subtasks under a generative perspective~\cite{chen-etal-2024-exploring-potential}. Representative works include using LLMs for extracting arguments~\cite{liu-etal-2023-argument,cabessa2024context,cabessa-etal-2025-argument,caputo-etal-2026-argument,elguendouze2026compactpromptinginstructiontunedllms} and relations~\cite{gorur-etal-2025-large} spanning multiple domains~\cite{pojoni2023argument, favero2025leveragingsmallllmsargument}.
Despite showing the promise of LLMs in AM, these works are often not designed for political debates, and do not extend argument mining with entity extraction.

Concerning the latter, LLMs have also been proven promising for Named Entity Recognition (NER) tasks. GPT-NER~\cite{wang-etal-2025-gpt} transforms NER from a supervised task to a text-generation one with self-verification to improve performance, PROMPT-NER~\cite{shen-etal-2023-promptner} unifies entity locating and entity typing in prompt learning for NER with a multi-prompt template, and Universal-NER~\cite{Zhou00CP24} distills large LLMs into smaller ones for open NER.
However, NER-oriented works do not model the argumentative layer required by AM scenarios.

Table~\ref{tab:related-comparison} positions our work in the current body of works addressing AM or generic NER, highlighting data- and task-related novelties and contributions, which allow us to fill a key gap in the literature.

\section{Problem Definition}
\label{sec:problem}

Let $D=[u_1,u_2,\ldots,u_n]$ be a debate transcript represented as an ordered sequence of turns. Each turn $u_i=(s_i,x_i)$ is a pair where $s_i$ denotes the speaker metadata, while $x_i=[t_{i_1},t_{i_2},\ldots,t_{m_i}]$  is a textual sequence, with each $t_{i_j}$ being a token drawn from a vocabulary $V$.

We consider a turn-based, streaming-compatible annotation setting. At time step $i$, the model observes only the current turn $u_i$ and must produce its annotations without access to previous turns $u_1,\ldots,u_{i-1}$ or future turns $u_{i+1},\ldots,u_n$.

\begin{definition}[Argument Component]
\label{def:argument-component}\rm
Given a turn $u_i$ with text $x_i=[t_{i_1},\ldots,t_{m_i}]$, an \textit{argument component} (AC) is a contiguous, non-overlapping span of tokens
$$
AC=[t_b,\ldots,t_e], \qquad i_1 \leq b \leq e \leq m_i,
$$
that expresses a meaningful argumentative unit. Each argument component is assigned an argumentative label $c$ from the set $
\mathcal{C}=\{\textsc{Claim},\textsc{Premise}\}.
$
A tagged argument component is obtained by enclosing the component span within paired tags corresponding to its argumentative label, i.e.,
$$
\langle c \rangle AC \langle /c \rangle, \qquad c \in \mathcal{C}.
$$
\end{definition}

\noindent\textbf{Example 1\ }
Given the turn \textit{``We must reduce taxes because families are struggling.''}, a possible argument-component annotation is: ``<\texttt{claim}>We must reduce taxes</\texttt{claim}> because <\texttt{premise}>families are struggling</\texttt{premise}>.''\hfill $\Box$

\begin{definition}[Debate Named Entity]
\label{def:named-entity}\rm
Given a turn  $u_i$ with text $x_i=[t_{i_1},\ldots,t_{m_i}]$, a \textit{debate named entity} (DNE) is a contiguous, non-overlapping span of tokens
$$
DNE=[t_b,\ldots,t_e], \qquad i_1 \leq b \leq e \leq m_i,
$$
that refers to a debate-relevant entity. Each named entity is assigned a type $\tau$ from a predefined set $\mathcal{T}$.
A tagged DNE is hence obtained by enclosing the entity span within paired tags with the corresponding entity type:
\texttt{<}$\tau$\texttt{>}NE\texttt{</}$\tau$\texttt{>}.
\end{definition}

We hereinafter refer to detecting and tagging DNE spans as \textit{Debate Named Entity Recognition} (DNER), i.e., a debate-specific NER task over the DNE inventory.
Note that DNEs can also be found, and hence tagged, outside argumentative spans.

\vspace{1.5mm}
\noindent\textbf{Example 2\ }
Given the turn ``President Biden spoke in Washington during the campaign.'', a possible named-entity annotation is:

``<\texttt{person}>President Biden<\texttt{/person}> spoke in <\texttt{location}>Washington<\texttt{/location}> during the <\texttt{event}>campaign</\texttt{event}>.''
\hfill $\Box$

\begin{definition}[Joint Argument--Entity Tagging]
\label{def:jaet}\rm
Given a turn $u_i$, \textit{Joint Argument--Entity Tagging} (JAET) is the task of learning a mapping function

$$
f(u_i)=u_i^*,
$$
where $u_i^*=(s_i,x_i^*)$, with $x_i^*$ being the annotated version of $x_i$ that preserves the original token order and inserts paired inline tags for argument components and debate-relevant named entities.

\end{definition}

\section{JAET Mapping Learning}
We formulate JAET as a supervised text-generation task, aimed at learning the transformation function $f$ from Definition~\ref{def:jaet}. We implement this formulation using a \textit{decoder-only} Large Language Model whose parameters $\theta$ are optimized via fine-tuning.

\begin{figure}[t]
    \centering
    \includegraphics[width=1\linewidth]{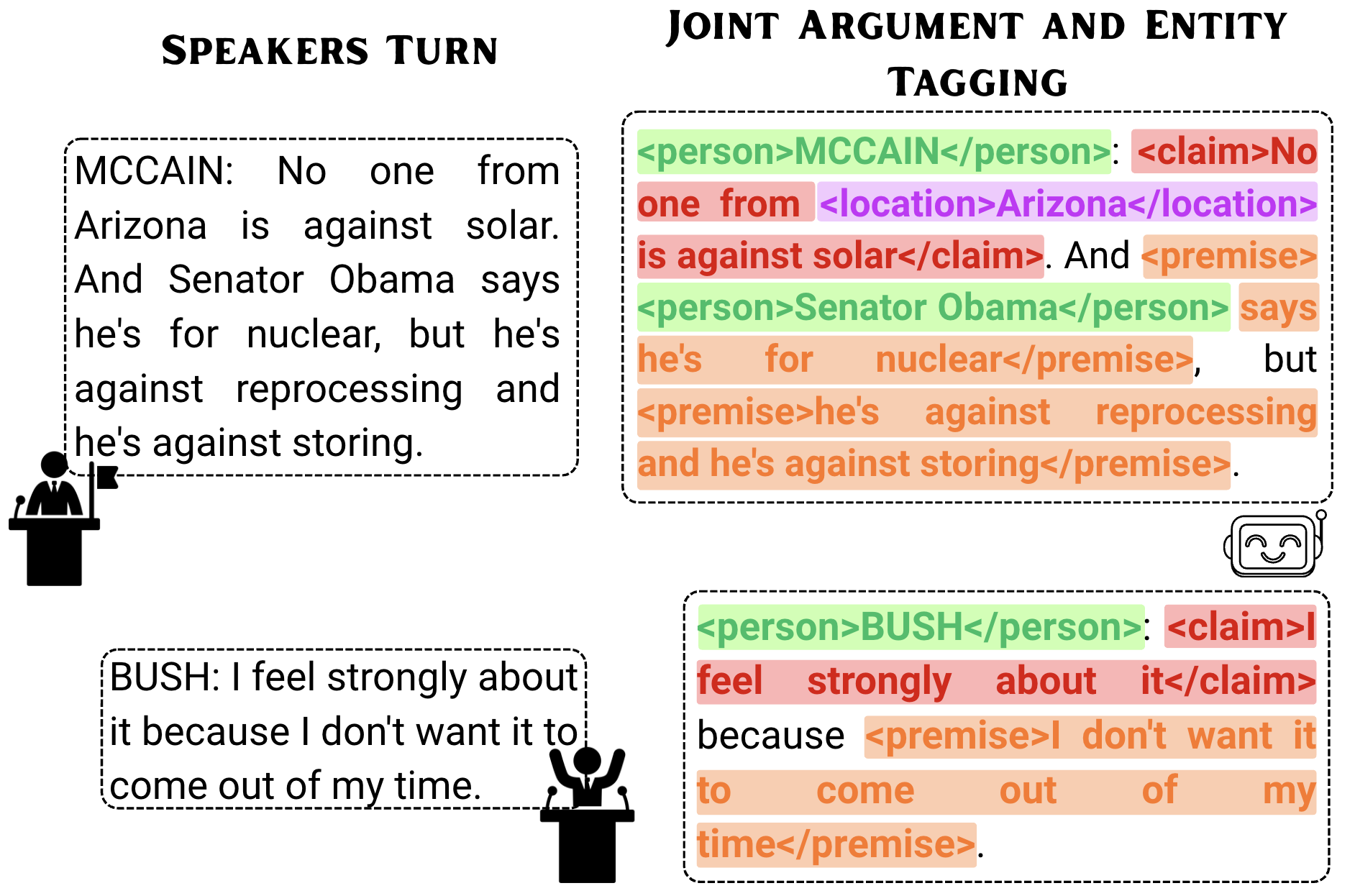}
    \caption{Example of JAET annotation on debate turns.}
    \label{fig:example-llm-annotation}
\end{figure}

\subsection{Learning Paradigm}
Each training instance corresponds to a debate turn $u_i = (s_i , x_i)$. 
The model must produce an output that (i) preserves the verbatim input text, and (ii) adds inline argument and entity tags when appropriate, as shown in Figure~\ref{fig:example-llm-annotation} (cf. Figure~\ref{fig:example-1} in Appendix~\ref{app:qualitative} for a detailed example).
To make our approach suitable for turn-based parsing of debates, no previous or future debate turns are included in the training units, forcing the model to produce annotations only relying on the turn's content.

\subsection{Learning Objective}
We start optimizing our model by creating a paired dataset $\{(D_i, D^\tau_i)\}_{i=1}^L$ containing $L$ debates $D_i=[u_{i,1}, u_{i,2},...,u_{i,n_i}]$ and corresponding ground-truth tagged versions $D^\tau_i = [u_{i,1}^{\tau}, u_{i,2}^{\tau},...,u_{i,n_i}^{\tau}]$, where $u_{i,j}$ indicates turn $j$ of debate $i$ and $\tau$ marks the tagged version.

Since our formulation operates at a turn-level, we unfold these debates into a supervised fine-tuning set according to the \textit{Alpaca} format, which has proven to be suitable for instruct fine-tuning of argument mining task-specific LLMs~\cite{liu-etal-2023-argument,cabessa-etal-2025-argument,caputo-etal-2026-argument}:
\begin{equation*}
    \mathcal{S}=\{(I,C_k,Y_k)\}_{k=1}^{N},
\end{equation*}
where $N=\sum_{i=1}^{L}n_i$ is the total number of flattened turns across all debates, $I$ is the single instruction applied to all input turns (cf. Figure~\ref{fig:instruction-sft} in Appendix~\ref{app:models}), $C_k=(s_k,x_k)$ is the input context containing speaker information and the raw text of flattened turn $k$, and $Y_k$ is the corresponding gold tagged output for that turn.

Accordingly, we optimize the set of parameters $\theta$ of the underlying decoder-only LLM by minimizing the negative log-likelihood of the tagged output. Specifically, we first consider a per-turn loss:
\begin{equation}
\!\!\!\!\!\!
\ell_k(\theta)\!
=\!
-\frac{1}{|Y_k|}\sum_{t=1}^{|Y_k|}
\log p_{\theta}\!\left(
y_{k,t}\mid y_{k,<t}, I, C_k
\right)
  \label{eq:turn-loss}
\end{equation}
where $y_{k,t}$ is the $t$-th token of $Y_k$. These values are hence aggregated over all turn-level examples as:
\begin{equation}
\mathcal{L}_{\text{JAET}}(\theta)
=
\frac{1}{N}
\sum_{k=1}^{N}
\ell_k(\theta).
  \label{eq:jaet-loss}
\end{equation}

Note that the normalization factor in Eq.~\ref{eq:turn-loss} prevents longer turns from dominating the optimization process, while Eq.~\ref{eq:jaet-loss} gives each turn-level sample the same weight. Furthermore, since each $Y_k$ contains the original turn tokens plus ground-truth tags, the model is penalized in case it (i) changes the original text, (ii) misses or hallucinates any tag, and (iii) assigns the wrong argument or entity label. Finally, since $C_k$ does not contain any previous or future turn, our training objective matches the desired turn-based inference setting.

\subsection{Models}
We consider \textit{small}, \textit{open-weight} Large Language Models that are publicly accessible through the Hugging Face Hub.
We deliberately focus on such models to support JAET adoption in AM scenarios that (i) require accessible and scalable deployment~\cite{favero2025leveragingsmallllmsargument}; (ii) exhibit low-resource constraints~\cite{kashefi-etal-2023-argument}; and (iii) require fine-tuning or deployment in privacy-preserving settings, e.g., legal~\cite{Habernal2023legalAM} or medical~\cite{mayer2020ecai} domains.

Following earlier work on AM~\cite{caputo-etal-2026-argument}, we resort to 7-8B parameters models, i.e., \textit{Llama 3.1 8B Instruct}, \textit{Mistral v0.3 7B Instruct}, and \textit{Qwen 2.5 7B Instruct}.
We report the full details on model deployment and fine-tuning in Appendix~\ref{app:models}.

\section{Data}
To train our set of LLMs and learn the JAET mapping, we introduce the tagged \dataset resource, which enriches the USElecDeb60to20 corpus, originally introduced by~\citet{haddadan-etal-2019-yes} and later updated by~\citet{goffredo-etal-2023-argument}, with debate-relevant entity annotations.
The original corpus consists of transcripts\footnote{Transcripts are publicly available from the Commission on Presidential Debates website at \url{debates.org}} from 44 television debates from the U.S. presidential and vice-presidential campaigns between 1960 and 2020, with a particular focus on reciprocal discussion between Democrat and Republican candidates. This captures the interactional structure of real-world political debates, and is therefore a natural and ideal setting for studying how arguments unfold around debate-relevant entities in turn-based political discourse.
We next describe how the inherited argument annotations were converted into inline tags and how the new entity layer was created.

\subsection{Argument Annotations}
We used the original human annotations from~\citet{haddadan-etal-2019-yes,goffredo-etal-2023-argument} as a starting point. To support generative inline tagging, we converted the inherited metadata annotations, which consist of a separate file containing only annotations, into paired typed tags surrounding the original text, i.e., \texttt{<claim>...</claim>} and \texttt{<premise>...</premise>}.
This conversion also required a transcript-alignment curation pass. We fixed minor incorrect or duplicate span positions, and restored original punctuation or wording that had been altered or omitted, when the inherited human annotations did not exactly match the original transcript.
All these operations ensured the human-annotated spans perfectly matched the original debate transcripts before adding entity tags.

\subsection{Entity Annotations}
\label{subsec:entity-annotations}
This layer is newly introduced in \dataset, and covers mentions to debate-relevant entities that are central to the interpretation of political debates, including entities inside argumentative spans as well as within contextual non-argumentative spans.

To define the debate entity inventory, we first combined manual inspection with LLM-assisted exploration of key political entities within the full set of debates considered in this study. We thus consolidated them into the following entity set:
$
\mathcal{T}=
\begin{aligned}[t]
\{&\textsc{Person},\textsc{Role},\textsc{Organization},\\
  &\textsc{Party},\textsc{Location},\textsc{Event},\textsc{Date}, \textsc{Law}\}.
\end{aligned}
$

We hence performed entity annotations and corresponding tag insertion within each debate through an LLM-assisted, human-validated workflow, which reduces the costs of human annotation, and follows prior work on LLM-assisted annotation~\cite{zhang-etal-2023-llmaaa,ehsan-solorio-2026-scalable}. 
For each turn, we first annotated Debate Named Entities using the Gemini APIs, prompting the model three times to annotate the text verbatim with the corresponding DNE tags, based on the approach illustrated in Appendix~\ref{app:annotation}. We then determined final entity tags through a consensus-based strategy: a tag is kept in the final annotation whenever the majority of runs assigned the same entity type to the same text span. 

The resulting annotations were validated as follows.
First, we quantitatively checked the consistency of LLM annotations across turns using the Fleiss' $\kappa$, which measures the degree of inter-annotator agreement beyond chance across multiple raters, at the entity-type level, obtaining $\kappa=0.724$. We treat this value only as a stability indicator of the repeated LLM-assisted annotation process, not as an inter-rater agreement measure. 

Second, we performed human validation by asking three domain experts to independently annotate a stratified sample of raw transcripts only, blind to LLM annotations. We achieved an almost perfect human agreement over the token-level BIO label sequences ($\kappa=0.919$). Remarkably, adding the Gemini consensus that produced the released annotations to the three experts confirmed the high agreement ($\kappa=0.900$), further strengthening our LLM-assisted annotation process.

Third, we also assessed the effect of introducing an additional frontier LLM-annotator, i.e., GPT-5.5, still obtaining a high agreement ($\kappa=0.862$). This suggests LLM-annotators tend to annotate at essentially the level of agreement the experts find among themselves. Interestingly, in Section~\ref{sec:robustness} we show that substituting the human annotation as ground truth leaves every result unchanged. 

We report the full annotation protocol, together with the recurring qualitative patterns during annotation and the systematic tendencies we observed in the LLM-assisted pipeline, in Appendix~\ref{app:annotation}.

\begin{table}[t!]
\centering
\small
\setlength{\tabcolsep}{3pt}
\resizebox{\linewidth}{!}{
\begin{tabular}{lccccc}
\toprule
\textbf{Split} & \textbf{Turns} & \textbf{Tagged Spans} & \textbf{Claim} & \textbf{Premise} & \textbf{Entities} \\
\midrule
Train & 6682 & 56,869 & 13,887 & 11,497 & 31,485 \\
Test & 1671 & 13,931 & 3402  & 2733 & 7796\\
\midrule
Total & 8353 & 70,800 & 17,289 & 14,230 & 39,281\\
\bottomrule
\end{tabular}
}
\caption{\dataset dataset statistics. Tagged spans include argument components and debate-relevant named entities. The split follows an 80/20 train/test partitioning, and was designed to prevent unbalanced distribution of argumentative spans across splits.\vspace*{-1.5mm}
}
\label{tab:dataset-stats}
\end{table}

\begin{table}[t!]
\centering
\setlength{\tabcolsep}{3pt}
\resizebox{\linewidth}{!}{
\begin{tabular}{lcccccccc}
\toprule
\textbf{Split} & \textsc{date} & \textsc{event} & \textsc{law} & \textsc{Org} &\textsc{Loc} & \textsc{party} & \textsc{person} & \textsc{role} \\
\midrule
Train & 2303 & 977 & 1098 & 3017 & 5939 & 773 & 13,625 & 3753 \\
Test & 545 & 219 & 266 & 765 & 1400 & 188 & 3438 & 975 \\
\midrule
Total & 2848 & 1196 & 1364 & 3782 & 7339 & 961 & 17,063 & 4728 \\
\bottomrule
\end{tabular}
}
\caption{Distribution of DNE tags across splits for each entity category (Org = Organizations, Loc = Locations).}
\label{tab:dataset-ner-stats}
\vspace{-5mm}
\end{table}

\vspace*{-1mm}
\subsection{Data Overview}
\label{sec:data-overview}
Table~\ref{tab:dataset-stats} summarizes the resulting dataset. The train/test split follows an 80/20 turn-level partition intentionally designed to ensure an argument-rich test set for properly evaluating AM and joint argument-entity tagging. This ensures a turn-level evaluation design for real-time debate analysis, rather than held-out-debate generalization.
Table~\ref{tab:dataset-ner-stats} reports the entity distribution across splits.

\vspace*{-1.5mm}
\section{Experimental Setup}
\label{sec:eval}
\vspace*{-1.5mm}
JAET is conceived as a joint generative tagging task. Accordingly, a correct output must (i) preserve the original turns, (ii) correctly identify and classify argumentative spans, and (iii) properly recognize political entities and their roles within debates. 

Let us denote with $\mathcal{D}=\{D_1, D_2, ..., D_L\}$ a collection of documents corresponding to $L$ political debates, such that each debate $D_i=[u_{i,1}, u_{i,2},...,u_{i,n_i}]$ consists of a list of $n_i$ turns.
We denote with $D_i^*$ the corresponding tagged version obtained by applying our LLM-based function $f$ to all turns in $D_i$, such that $D_i^*=[u_{i,1}^*, u_{i,2}^*,...,u_{i,n_i}^*]$, where each $u_{i,j}^*=f(u_{i,j})$.

Hereinafter, to keep the metric definition readable, we use $u$ to denote an arbitrary turn from the test set $U=\{u_{i,j} \, | \, D_i \in \mathcal{D} \, \wedge \, 1 \leq j \leq n_i \}$, and $u^*$ and $u^\tau$, to refer to the predicted and gold-tagged version of $u$, respectively.
Joint annotations are evaluated at the token level by pairing the argument and entity labels produced for the same token.

\subsection{Baselines and Competing Methods}
We compare the tagging and corresponding classification performance of JAET against two families of approaches, namely (i) prompt-based baselines, and (ii) task-specific competing methods. Specifically, the former allows us to quantify how far instruction-following alone can handle joint AM and DNER without any parameter update, while the latter allows us to compare JAET with the closest architectural alternatives. 
We report all details, prompts, and training settings for the baselines and competing methods presented next in Appendix~\ref{app:models}.

\paragraph{Prompt-based Baselines}
We consider zero-shot and few-shot settings adopting the same decoder-only LLMs used for JAET fine-tuning, to isolate the impact of the latter on base models. In the zero-shot setting, we provide the model only with task instruction, the current turn, the argument and entity type inventories, and the constraint to preserve original text. For few-shot, we also provide a small set of tagged turns sampled from the training split. In both cases, the LLM is asked to produce both argument and entity annotations.

\paragraph{Decoder-only Approaches}
As a generative competing method, we apply the set of models released by~\citet{caputo-etal-2026-argument} to \dataset. This approach uses the same family of 7-8B models we adopt in this work, thus allowing us to assess the impact of our tagging and fine-tuning strategy, but it is designed only for AM tasks and does not address DNER. Nonetheless, it currently represents one of the most competitive approaches for AM, and therefore constitutes a strong comparison point for the AM component of our framework.

\paragraph{Encoder-only Approaches}
We consider the recently released RooseBERT~\cite{RooseBERT}. This represents the closest approach w.r.t. our work, as it is a political-domain BERT-based model, which has been tested under both AM and entity-recognition scenarios, producing notable results.

To keep the evaluation fair, we fine-tuned this model to perform joint AM and DNER on the same \dataset train split used for JAET, using the parameters recommended by~\citet{RooseBERT}.

\subsection{Evaluation Criteria}
To evaluate the tagging and classification quality of JAET and competing methods, we focus on token-level segmentation quality, following standard evaluation practices in AM~\cite{RooseBERT,favero2025leveragingsmallllmsargument,caputo-etal-2026-argument}. We convert tagged turns into BIO sequences~\cite{ramshaw-marcus-1995-text} under two variants. In the \textit{boundary-only} variant, tokens are labeled as \textsc{B}, \textsc{I}, or \textsc{O} depending on whether they begin a span, continue a span, or occur outside any target span, regardless of the span type. In the \textit{typed} variant, boundary labels are paired with the corresponding class, e.g., \textsc{B-Claim}/\textsc{I-Claim} for argument components or \textsc{B-Person}/\textsc{I-Person} for entities, while \textsc{O} still denotes tokens outside the target layer.
For the joint setting, each token receives a paired label from both AM and debate-entity layers, and a prediction is considered correct iff both layers match the ground-truth label for that token.

For all scenarios, we report precision, recall, and $F_1$ computed over aligned BIO tags. 
Note that, to account for the imbalance in the set of argument classes and political entities, all scores are reported as macro-averaged.

Since token-level BIO evaluation requires predicted and gold sequences to be aligned, we enforce equal-length BIO sequences by right-padding the shorter sequence with \textsc{O} tags, enabling direct token-wise comparison under a one to one alignment strategy that penalizes generated outputs that insert, delete, or reorder transcript tokens.

\begin{table*}[t!]
\centering
\small
\setlength{\tabcolsep}{4pt}
\resizebox{\linewidth}{!}{
\begin{tabular}{llcccccccccccc}
\toprule
\multirow{2}{*}{\textbf{Approach}} &
\multirow{2}{*}{\textbf{Model}} &
\multicolumn{3}{c}{\textbf{Arg. Min.}} &
\multicolumn{3}{c}{\textbf{Debate NER}} &
\multicolumn{3}{c}{\textbf{Joint}} &
\multicolumn{2}{c}{\textbf{Sanity}} \\

\cmidrule(lr){3-5}\cmidrule(lr){6-8}\cmidrule(lr){9-11}\cmidrule(lr){12-13}

 & & P & R & $F_1$
 & P & R & $F_1$
 & P & R & $F_1$
 & $T_{\mathrm{PR}}$ & $T_{\mathrm{WR}}$  \\
\midrule
\multirow{3}{*}{\textbf{Ours}}
& Llama 3.1 8B 
& 0.741 & 0.740 & 0.740
& \textbf{0.927} & \textbf{0.909} & \textbf{0.918} 
& 0.643 & 0.635 & 0.638
& 0.955 & \textbf{0.999}  \\

& Mistral 7B 
& \textbf{0.749} & \textbf{0.742} & \textbf{0.745} 
& 0.919 & 0.904 & 0.912 
& \textbf{0.650} & \textbf{0.638} & \textbf{0.644}
& \textbf{0.959} & 0.998  \\

& Qwen 2.5 7B 
& 0.712 & 0.704 & 0.708 
& 0.920 & 0.898 & 0.909
& 0.623 & 0.610 & 0.616
& 0.946 & \textbf{0.999}  \\

\midrule
\multirow{1}{*}{\citet{RooseBERT}}
& RooseBERT 
& 0.535 & 0.539 & 0.530 
& 0.630 & 0.595 & 0.610 
& 0.365 & 0.397 & 0.344
& 0.587 & 0.948  \\
\midrule
\multirow{3}{*}{\citet{caputo-etal-2026-argument}}
& Llama 3.1 8B 
& 0.484 & 0.523 & 0.479 
& -- & -- & -- 
& -- & -- & -- 
& 0.473 & 0.950  \\

& Mistral 7B 
& 0.435 & 0.478 & 0.436 
& -- & -- & -- 
& -- & -- & --
& 0.586 & 0.955  \\

& Qwen 2.5 7B 
& 0.625 & 0.584 & 0.601 
& -- & -- & -- 
& -- & -- & -- 
& 0.834 & 0.980  \\
\midrule
\multirow{3}{*}{Zero-shot}
& Llama 3.1 8B 
& 0.361 & 0.348 & 0.315 
& 0.458 & 0.517 & 0.479 
& 0.161 & 0.183 & 0.146
& 0.104  & 0.793  \\

& Mistral 7B 
& 0.423 & 0.420 & 0.396 
& 0.351 & 0.345 & 0.346 
& 0.146 & 0.146 & 0.133 
& 0.008 & 0.789 \\

& Qwen 2.5 7B 
& 0.443 & 0.435 & 0.438 
& 0.365 & 0.364 & 0.364 
& 0.174 & 0.156 & 0.151
& 0.414 & 0.839 \\

\midrule
\multirow{3}{*}{Few-shot}
& Llama 3.1 8B 
& 0.523 & 0.547 & 0.532 
& 0.593 & 0.670 & 0.624 
& 0.320 & 0.347 & 0.317
& 0.522 & 0.889  \\

& Mistral 7B 
& 0.430 & 0.417 & 0.419 
& 0.661 & 0.563 & 0.602 
& 0.275 & 0.236 & 0.233
& 0.436 & 0.872  \\

& Qwen 2.5 7B 
& 0.464 & 0.386 & 0.356 
& 0.790 & 0.608 & 0.673
& 0.339 & 0.245 & 0.234
& 0.654 & 0.969  \\
\bottomrule
\end{tabular}
}
\caption{Boundary-only token-level BIO results on the test set of \dataset. Argument Mining (i.e., joint Argument Component Segmentation and Classification), Debate NER, and Joint report precision (P), recall (R), and $F_1$. $T_{PR}$/$T_{WR}$ for the encoder-only approach are obtained by reconstructing the text after annotation. ``--'' indicates unavailable outputs from that method. Bolded values correspond to the best performance. \vspace{-2mm}}
\label{tab:main-results}
\end{table*}

\begin{table*}[t!]
\centering
\setlength{\tabcolsep}{3pt}
\resizebox{\linewidth}{!}{
\begin{tabular}{llccccccccccccccc}
\toprule
\multirow{2}{*}{\textbf{Approach}} 
& \multirow{2}{*}{\textbf{Model}}
& \multicolumn{3}{c}{\textbf{B-Claim}}
& \multicolumn{3}{c}{\textbf{B-Premise}}
& \multicolumn{3}{c}{\textbf{I-Claim}}
& \multicolumn{3}{c}{\textbf{I-Premise}}
& \multicolumn{3}{c}{\textbf{O}} \\

\cmidrule(lr){3-5}
\cmidrule(lr){6-8}
\cmidrule(lr){9-11}
\cmidrule(lr){12-14}
\cmidrule(lr){15-17}

& 
& P & R & $F_1$
& P & R & $F_1$
& P & R & $F_1$
& P & R & $F_1$
& P & R & $F_1$ \\

\midrule

\multirow{3}{*}{\textbf{Ours}}
& Llama 3.1 8B
& 0.516 & 0.534 & 0.525
& 0.452 & 0.444 & 0.448
& \textbf{0.583} & 0.630 & 0.605
& 0.540 & 0.557 & 0.548
& 0.764 & 0.706 & 0.734 \\

& Mistral 7B
& \textbf{0.539} & \textbf{0.541} & \textbf{0.540}
& \textbf{0.476} & \textbf{0.456} & \textbf{0.466}
& 0.580 & \textbf{0.636} & \textbf{0.607}
& \textbf{0.560} & \textbf{0.576} & \textbf{0.568}
& 0.770 & 0.709 & 0.738 \\

& Qwen 2.5 7B
& 0.514 & 0.502 & 0.508
& 0.442 & 0.416 & 0.429
& 0.580 & 0.590 & 0.585
& 0.526 & 0.543 & 0.535
& 0.711 & 0.691 & 0.701 \\

\midrule

\citet{RooseBERT}
& RooseBERT
& 0.087 & 0.159 & 0.113
& 0.056 & 0.106 & 0.073
& 0.517 & 0.552 & 0.534
& 0.522 & 0.548 & 0.535
& 0.777 & 0.653 & 0.710 \\

\midrule

\multirow{3}{*}{Zero-shot}
& Llama 3.1 8B
& 0.036 & 0.028 & 0.032
& 0.054 & 0.016 & 0.024
& 0.306 & 0.145 & 0.197
& 0.260 & 0.062 & 0.100
& 0.511 & \textbf{0.826} & 0.631 \\

& Mistral 7B
& 0.019 & 0.021 & 0.020
& 0.019 & 0.029 & 0.023
& 0.246 & 0.370 & 0.296
& 0.265 & 0.481 & 0.342
& 0.772 & 0.454 & 0.572 \\

& Qwen 2.5 7B
& 0.083 & 0.074 & 0.079
& 0.059 & 0.031 & 0.041
& 0.338 & 0.447 & 0.385
& 0.340 & 0.254 & 0.291
& 0.585 & 0.570 & 0.577 \\

\midrule

\multirow{3}{*}{Few-shot}
& Llama 3.1 8B
& 0.131 & 0.127 & 0.129
& 0.124 & 0.194 & 0.151
& 0.322 & 0.226 & 0.266
& 0.305 & 0.513 & 0.383
& \textbf{0.816} & 0.744 & \textbf{0.778} \\

& Mistral 7B
& 0.134 & 0.121 & 0.127
& 0.148 & 0.103 & 0.122
& 0.330 & 0.268 & 0.295
& 0.377 & 0.322 & 0.347
& 0.468 & 0.574 & 0.516 \\

& Qwen 2.5 7B
& 0.193 & 0.076 & 0.109
& 0.237 & 0.055 & 0.090
& 0.359 & 0.162 & 0.223
& 0.396 & 0.116 & 0.180
& 0.432 & 0.816 & 0.565\\

\bottomrule
\end{tabular}
}
\caption{Typed token-level BIO results for AM (joint ACS and ACC) on the test set of \dataset. \citet{caputo-etal-2026-argument} does not produce joint ACS+ACC labels. Bolded values correspond to the best performance.\vspace{-3.5mm}}
\label{tab:main-results-details}
\end{table*}

\subsection{Sanity Check Criteria}
Finally, since JAET performs generative inline tagging, a tagged debate can be semantically valid but practically unusable, e.g., if the source text is changed, the LLM hallucinates, or tags are malformed. Therefore, we measure these as follows. 

Let us denote with $strip(\cdot)$ a cleaning function that removes all tags from an input sequence. 

To detect changes in tagged texts with respect to the original ones (e.g., due to hallucinations), we define the \textit{Text Preservation Rate} as:
\begin{equation}
\mathrm{T_{PR}} =
\frac{1}{|\mathcal{U}|}\sum_{u\in\mathcal{U}}
\mathbf{1}\left[
\text{strip}(u^*)={\text{strip}(u^\tau)}
\right].
\end{equation}

Similarly, we quantify tag syntax quality by means of the \textit{Tag Well-formed Rate} as:
\begin{equation}
\mathrm{T_{WR}} =
\frac{1}{|\mathcal{U}|}\sum_{u\in\mathcal{U}}
\mathbf{1}\left[
\text{wellformed}(u^*)
\right],
\end{equation}
where $\text{wellformed}(\cdot)$ checks whether tags are properly formatted, opened, closed, and nested.

\section{Results}
\label{sec:results}
\vspace*{-2mm}
Table~\ref{tab:main-results} reports boundary-only token-level BIO results for argument mining,  debate entity recognition, and the joint task. 
Across all scenarios, JAET variants outperform the competing methods and prompt-based baselines: Mistral 7B emerges as the strongest variant for AM ($F_1=0.745$) and joint AM+DNER ($F_1=0.644$)---becoming our reference model, while Llama 3.1 8B achieves the highest debate-entity score ($F_1=0.918$). 

Specifically, comparing each task against the strongest non-JAET baseline, JAET improves relative $F_1$ by +24\% for AM over the best AM-only model by~\citet{caputo-etal-2026-argument} (0.745 vs. 0.601), +36.4\% for DNER over the best prompt-only baseline (0.918 vs. 0.673), and +87.2\% for joint AM-DNER over RooseBERT (0.644 vs. 0.344). 
Note that the latter is conceived as an encoder-only approach, and this is reflected in generative-quality aspects, e.g., its low text preservation rate. Therefore, while keeping it within our comparison, we consider models having the same generative backbone as direct competing approaches.

Prompt-only baselines clarify the role of fine-tuning, showing that simple prompting is not sufficient for reliable joint tagging. 
Indeed, while few-shot prompting improves over zero-shot for AM (Llama $F_1= 0.532$), DNER (Qwen $F_1= 0.673$), and joint tagging (Llama $F_1=0.317$), suggesting provided examples help models understanding the desired schema, all remain far below the best JAET scores of 0.745, 0.918, and 0.644, respectively.

The sanity metrics from Table~\ref{tab:main-results} further show these gains are not obtained at the cost of malformed or heavily altered outputs, since JAET consistently yields text preservation rate at least 0.946 and tag well-formed rate at least 0.998. By contrast, the best non-JAET text preservation rate is 0.834, and prompt-only variants are even less reliable in preserving the original script. 
Furthermore, repeating fine-tuning and inference under three random seeds makes macro-$F_1$ vary below $10^{-3}$, thus the reported improvements are outside seed variance.

Note that $T_{PR}$ and $F_1$ are not fully orthogonal, since our BIO alignment right-pads the shorter sequence (cf. Section~\ref{sec:eval}). Considering correctly-preserved turns only (cf. Table~\ref{tab:preserved} in Appendix~\ref{app:results}) still leaves a gap of 0.19 $F_1$ on AM and 0.26 on the joint task in favor of JAET, which is hence due to tagging quality rather than text preservation.  

Given the high DNE capabilities of all considered approaches, we next focus on typed BIO AM results, unveiling where potential AM bottlenecks emerge. As reported in Table~\ref{tab:main-results-details}, the beginning of argumentative spans represents the major bottleneck for AM, with \textsc{B-*} labels being consistently harder than the corresponding \textsc{I-*} ones. 

To isolate the source of the JAET gains over competing methods, we compared the best-performing JAET model (i.e., Mistral) with three ablated variants that share the same set of models and fine-tuning hyperparameters than the full JAET one. AM-only removes the entity layer from the training data, and is meant to assess whether and to what extent joint tagging harms AM performance. AM$\to$DNER and DNER$\to$AM represent sequential pipeline variants in which the two layers are predicted in a fixed order rather than jointly, by means of task-specific models.

Ablation results in Table~\ref{tab:ablation-combined} confirm that the remarkable performance achieved by JAET are due to joint argument-entity modeling. 
Interestingly, compared with the AM-only variant, JAET fully preserves the AM performance under both untyped and typed evaluation settings: recovering the entity layer and the joint structure comes at no cost on argument mining, which is a desiderata of the entity-enriched formulation we propose.

Most importantly, Table~\ref{tab:ablation-combined} demonstrates that JAET performance cannot be achieved by simply leveraging a sequential AM-DNER or DNER-AM pipeline. Indeed, JAET improves by +27.3\% relative $F_1$ over the best sequential pipeline (AM$\to$DNER, 0.644 vs. 0.506) considering the untyped scenario, and by +41.9\% over the typed one (0.464 vs. 0.327). Against DNER$\to$AM the corresponding margins increase to +55.2\% and +122\%, respectively.
The latter suggests that JAET learns better tag placement, while also improving type inference, compared to pipeline-based variants. 
Appendix~\ref{app:qualitative} reports additional insights into the underlying error patterns of sequential composition that a single-pass model cannot exhibit by construction.
Overall, our results support the central claim that predicting arguments and entities in a single pass better exploits their interdependency than composing two independent modules.

\begin{table}[t!]
\centering
\small
\setlength{\tabcolsep}{2.5pt}
\begin{tabular}{clcccccc}
\toprule
&
\multirow{2}{*}{\textbf{Approach}} &
\multicolumn{3}{c}{\textbf{Arg. Min.}} &
\multicolumn{3}{c}{\textbf{Joint}} \\

\cmidrule(lr){3-5}\cmidrule(lr){6-8}

& & P & R & $F_1$
& P & R & $F_1$ \\
\midrule

\multirow{4}{*}{\rotatebox{90}{Untyped}}
& Our
& \textbf{0.749} & \textbf{0.742} & \textbf{0.745} 
& \textbf{0.650} & \textbf{0.638} & \textbf{0.644} \\

& AM-only
& 0.741 & 0.731 & 0.736
& -- & -- & -- \\

& AM$\rightarrow$DNER
& 0.716 & 0.694 & 0.703
& 0.633 & 0.498 & 0.506 \\

& DNER$\rightarrow$AM
& 0.705 & 0.682 & 0.690
& 0.516 & 0.409 & 0.415 \\

\midrule

\multirow{4}{*}{\rotatebox{90}{Typed}}
& Our
& 0.585 & \textbf{0.584} & \textbf{0.584} 
& \textbf{0.489} & \textbf{0.461} & \textbf{0.464} \\

& AM-only
& \textbf{0.588} & 0.576 & 0.582
& -- & -- & -- \\

& AM$\rightarrow$DNER
& 0.567 & 0.539 & 0.551
& 0.382 & 0.330 & 0.327 \\

& DNER$\rightarrow$AM
& 0.558 & 0.528 & 0.540
& 0.361 & 0.196 & 0.209 \\

\bottomrule
\end{tabular}
\caption{Ablation study comparing Mistral JAET against AM-DNER pipelines and single-task variants on Argument Mining (AM) under typed and untyped BIO settings. Bolded values indicate the best performance.\vspace{-3mm}}
\label{tab:ablation-combined}
\end{table}

\section{Robustness and Generalization}
\label{sec:robustness}
We finally assess the robustness of our findings using our reference model Mistral, as detailed next.

\vspace*{-1mm}
\paragraph{Annotation Provenance}
First, we verify that our results do not depend on the LLM-assisted entity layer annotations (cf. Section~\ref{subsec:entity-annotations}) by replacing these annotations with those produced by three human experts and replicating our experiments. Notably, we do not observe any concrete change in $F_1$ (e.g., joint untyped $F_1$ moves from 0.6436 to 0.6448), and the joint scores slightly increase. This confirms JAET performance is invariant to whether the ground truth is model- or human-annotated.

\vspace*{-1mm}
\paragraph{Generalization to Unseen Debates}
By default, our data partition is turn-level to target real-time debate analysis rather than fully held-out-debates (cf. Section~\ref{sec:data-overview}). However, to discard any potential leakage and measure how well our approach generalized to unseen debates, we re-partitioned our data at the debate level, i.e., 35 debates for training and 9 entirely held-out ones for testing. We hence re-trained our reference model and repeated the evaluation. Notably, this process only yielded a cost of 2.4--5.0\% relative $F_1$ (cf. Table~\ref{tab:robustness} in Appendix~\ref{app:robustness}). In particular, typed AM precision increased on held-out debates (0.5851 vs. 0.5945) while recall decreases (0.5835 vs. 0.5576), i.e., the model becomes more conservative on unfamiliar spans, which we ascribe to genuine generalization rather than memorization.

\paragraph{Domain Transfer}
We finally assess whether our findings are domain-specific, by evaluating JAET on Persuasive Essays~\cite{stab2017persuasiveessay}, a domain differing from political debates on every relevant axis: written rather than spoken, single-author rather than adversarial, and edited rather than disfluent. 
Our reference model achieves untyped $F_1=0.8641$ on AM and $0.6844$ on the joint task, improving over the strongest prompt-based baseline by +104\%, resp. 235.5\% (cf. Table~\ref{tab:pe-full} in Appendix~\ref{app:robustness}). Most importantly, repeating the full ablation in this domain replicates our central finding (cf. Table~\ref{tab:pe-ablation}): JAET improves over the strongest sequential pipeline---DNER$\to$AM here---by +26.6\% untyped and +52.7\% typed, closely mirroring the debate margins (+27.3\%, resp. +41.9\%). Exactly as in debates, joint tagging matches a dedicated AM-only model on AM while additionally recovering the entity layer a pipeline cannot produce, and also properly generalizes across domains.

\section{Conclusions}
In this work, we made a two-fold contribution to political debate analysis. First, we curated \dataset, an entity-enriched extension of the USElecDeb dataset which addresses the need of unveiling key entities involved in political debates for improving argument mining. Second, we proposed JAET, a single-pass generative framework for joint argument mining and debate named entity recognition.
A thorough experimental evaluation showed that JAET enables effective online tag insertion, outperforming encoder-only and decoder-only methods as well as AM-DNER sequential pipelines, with gains that cannot be recovered by composing independent modules and that hold even under strict robustness and generalization assessments.

Future work will investigate extensions to argument and entity relations, and richer cross-turn dependencies, key ingredients for enabling contextually-rich, structured representations of political and debate argumentation.

\vspace{1mm}
\noindent All code, models, and resources associated with this work are publicly available at \url{https://github.com/stemonea/JAET}.

\vspace*{-1mm}
\paragraph{Acknowledgments.}
The paper was partially supported by the MUR PRIN~2022 project S-PIC4CHU (2022XERWK9).

\section*{Limitations}

\noindent\textbf{Language Use}
\dataset is built from U.S. presidential and vice-presidential debate transcripts, which correspond to English-only texts. We acknowledge certain discourse patterns may not transfer to other languages or cultures, and therefore broadening the set of languages remains an open direction.

\vspace{1.5mm}
\noindent\textbf{Entity Inventory}
The entity inventory used in this work is extracted from U.S. presidential and vice-presidential debate transcripts. Certain roles, organizations, parties, and discourse patterns might not generalize to other political scenarios or cultures. Enhancing generalizability to other settings (e.g., non-Western) remains an open challenge.

\vspace{1.5mm}
\noindent\textbf{Turn-level Perspective}
Our framework operates only on a turn-level perspective for two main reasons. First, aggregating whole, hour-long debates goes beyond the context window of the small, deployable models we target. Second, a preliminary evaluation of our reference model on inputs aggregating multiple consecutive turns was found to affect every metric and collapse text preservation (cf. Table~\ref{tab:window} in Appendix~\ref{app:robustness}). 
Nonetheless, while turn-level operation is an optimal operating setting for joint tagging, we acknowledge it limits the detection of cross-talks and co-references, which represent a natural direction for improvement. 

\vspace{1.5mm}
\noindent\textbf{Annotation Provenance}
The entity layer of \dataset is produced through an LLM-assisted, human-validated workflow rather than by exhaustive human annotation. While we emphasize that human experts report almost perfect agreement (cf. Section~\ref{subsec:entity-annotations}) and using only the human annotations as ground truth leaves every result unchanged (cf. Section~\ref{sec:robustness}), we acknowledge the annotation pipeline might exhibit some systematic tendencies, which we report in Appendix~\ref{app:annotation}.

\section*{Ethical Considerations}
The enriched dataset and novel framework we release in this work might be used for automated political debate analysis. Notably, the latter might still affect how candidates, parties, and public debates are interpreted if used without any human supervision. We therefore urge all stakeholders to use our, and similar tools as an analytical aid rather than a decision-making tool, and we discard any misuse or decisions taken using our work.

\appendix

\section{Implementation Details}
\label{app:models}
All experiments were conducted on a fixed hardware setup consisting of 2$\times$NVIDIA Tesla T4 GPUs, each equipped with 15 GB of VRAM. 

\begin{table}[ht!]
\centering
\small
\setlength{\tabcolsep}{5pt}
\begin{tabular}{ll}
\toprule
\textbf{Model} & \textbf{Hugging Face ID} \\
\midrule
Llama 3.1 8B    & meta-llama-3.1-8b-instruct-bnb-4bit \\
Mistral v0.3 7B & mistral-7b-instruct-v0.3-bnb-4bit \\
Qwen 2.5 7B     & Qwen2.5-7B-Instruct-unsloth-bnb-4bit \\
\bottomrule
\end{tabular}
\caption{Hugging Face model identifiers for the pre-trained checkpoints used in our experiments.}
\label{tab:model-id}
\end{table}

\begin{table}[ht!]
\centering
\small
\renewcommand{\arraystretch}{1.15}
\begin{tabular}{llr}
\toprule
\textbf{Category} & \textbf{Parameter} & \textbf{Value} \\
\midrule
\multirow{5}{*}{Training Setup}
& Epochs & 3 \\
& Batch size & 2 \\
& Grad. accum. steps & 4 \\
& Effective batch size & 8 \\
& Max. seq. length & 4096 \\
\midrule
\multirow{5}{*}{Optimization}
& Learning rate & $1 \times 10^{-4}$ \\
& Weight decay & $1 \times 10^{-3}$ \\
& LR scheduler & Linear \\
& Warmup steps & 5 \\
& Optimizer & AdamW (8-bit) \\
\midrule
\multirow{4}{*}{LoRA Config}
& Fine-tuning type & LoRA \\
& Target modules & All linear layers \\
& Rank ($r$) & 16 \\
& Scaling ($\alpha$) & 16 \\
& Dropout & 0 \\
& Bias & None \\
\midrule
Quantization & Precision  & 4-bit (NF4) \\
\midrule
\multirow{2}{*}{Inference Setup}
& Max.\ new tokens & 2048 \\
& Temperature & $0.01$ \\
& Do-Sample & true \\
& top-$p$ & 0.1\\
\bottomrule
\end{tabular}
\caption{Hyperparameter configuration for supervised fine-tuning (SFT) via Unsloth. The inference hyperparameters were applied uniformly across
all experimental conditions, including zero/few-shot prompting settings.}
\label{tab:hyperparameters-fine-tuning}
\end{table}

Concerning decoder-only models, Table~\ref{tab:model-id} lists the Hugging Face model identifiers used to load the pre-trained checkpoints, while Table~\ref{tab:hyperparameters-fine-tuning} summarizes the full set of hyperparameters adopted during both training and inference phases. 
To mitigate memory constraints while preserving robust model performance, all models were fine-tuned using parameter-efficient methods, specifically Low-Rank Adaptation (LoRA) via the Unsloth framework, combined with 4-bit quantization. The optimization procedure was kept consistent across all architectures to ensure fair comparability of results. 
Training data was formatted following the Alpaca-style instruction format, as illustrated in Figure~\ref{fig:alpaca-formatting}. The \texttt{instruction} field contains the full task prompt (Figure~\ref{fig:instruction-sft}), which specifies the annotation schema and constraints the model must follow during generation. The inference hyperparameters reported in Table~\ref{tab:hyperparameters-fine-tuning} were applied uniformly across all experimental conditions, including zero-shot and few-shot prompting settings.
Finally, concerning the RooseBERT encoder-only approach, we trained it for 3 epochs, with batch size 16 and max sequence length of 512, using a learning rate of $3\times10^{-5}$ and weight decay of 0.01.

\begin{figure}[t!]
\centering
\begin{tcolorbox}[
    width=\columnwidth,
    colback=white,
    colframe=black!80,
    coltitle=white,
    fonttitle=\bfseries,
    title={Alpaca Formatting},
    boxrule=1pt,
    arc=2mm,
    top=2mm,
    bottom=2mm,
]
\footnotesize
$\{$"\textbf{instruction}": "You are given a transcript of a political debate. Your task is to annotate the text by identifying ... ",\\
"\textbf{input}": "MR. McGEE: Senator Kennedy, would you ...":\\
"\textbf{output}": "<person>MR. McGEE</person>: <person>Senator Kennedy</person> ..."$\}$
\end{tcolorbox}
\caption{Example of a training instance formatted according to the Alpaca-style instruction
template.}
\label{fig:alpaca-formatting}
\end{figure}

\begin{figure}[t!]
\centering
\begin{tcolorbox}[
    width=\columnwidth,
    colback=white,
    colframe=black!80,
    coltitle=white,
    fonttitle=\bfseries,
    title={Full Instruction},
    boxrule=1pt,
    arc=2mm,
    top=2mm,
    bottom=2mm,
]
\footnotesize
You are given a transcript of a political debate. Your task is to annotate the text by identifying both argumentative components and named entities. You must preserve the original
text exactly and insert XML-style tags directly into it without altering, reordering, or paraphrasing any words. For argumentative structure, identify claims and premises. A claim
is a statement that expresses a position, opinion, or conclusion. A premise is a statement that provides evidence, justification, or reasoning supporting a claim. Wrap claims with
\texttt{<claim>...</claim>} and premises with \texttt{<premise>...</premise>}.
For named entities, identify and annotate all occurrences of persons, organizations, locations, and roles. Use the following tags: \texttt{<person>...</person>}, \texttt{<role>...</role>}, \texttt{<organization>...</organization>}, \texttt{<party>...</party>}, \texttt{<location>...</location>}, \texttt{<event>...</event>}, \texttt{<date>...</date>}, and \texttt{<law>...</law>}. Annotations must be applied directly to the text so
that all tags are properly nested and do not overlap incorrectly. If a named entity appears inside a claim or premise, the entity tag must be fully contained within the argument tag. 
Do not create crossing or partially overlapping tags. Do not annotate text that is not part of an argument as a claim or premise. If a sentence does not contain argumentative content,
leave it unchanged except for possible named entity annotations. Return only the fully annotated text and nothing else.
\end{tcolorbox}
\caption{Full instruction prompt used during supervised fine-tuning and inference.}
\label{fig:instruction-sft}

\end{figure}

\section{Details on Entity Annotation}
\label{app:annotation}

\paragraph{LLM-based Annotations}
For entity annotation, we leveraged \textit{gemini-3.1-pro-preview} through the Google AI Studio API interface\footnote{\url{https://aistudio.google.com/}} with temperature 1.0, top-p set to 0.95, and thinking level set to high.
These settings were selected to balance annotation consistency and contextual sensitivity during entity extraction from political debate transcripts. 
Figure~\ref{fig:prompt_ner} reports the exact prompt template used to instruct the model during the annotation process.

\begin{figure}[t!]
\centering

\begin{tcolorbox}[
    width=\columnwidth,
    colback=white,
    colframe=black!80,
    coltitle=white,
    fonttitle=\bfseries,
    title={Prompt for Entity Annotation},
    boxrule=1pt,
    arc=2mm,
    top=2mm,
    bottom=2mm,
]
\footnotesize

\textit{You are an expert linguistic annotator. Your task is to perform semantic segmentation and named entity recognition (NER) on the given text.}

\vspace{1mm}
\textbf{TAGGING SCHEME}

\begin{itemize}\setlength\itemsep{1pt}
    \item \texttt{<person>}: Names of individuals, including references with titles, honorifics, or descriptive modifiers.
    \item \texttt{<organization>}: Institutions, agencies, committees, companies, media outlets, and political organizations.
    \item \texttt{<party>}: Political parties, coalitions, or political movements.
    \item \texttt{<location>}: Geographic entities such as countries, cities, states, or regions.
    \item \texttt{<role>}: Institutional positions or occupations when used generically.
    \item \texttt{<event>}: Political, historical, or public events, including elections, debates, and campaigns.
    \item \texttt{<date>}: Explicit temporal expressions such as years, dates, or months.
    \item \texttt{<law>}: Named laws, acts, constitutional amendments, or legal provisions.
\end{itemize}

\textbf{ANNOTATION CONSTRAINTS}

\begin{enumerate}\setlength\itemsep{1pt}
    \item Do not modify the original text in any way.
    \item Only insert tags around existing text spans.
    \item Annotate the complete entity span whenever possible.
    \item Use \texttt{<role>} for generic institutional references.
    \item Do not introduce nested tags.
    \item Do not remove or reorder any content.
    \item Return the original text with tags only.
\end{enumerate}

\textbf{EXAMPLE}

\vspace{1mm}
\texttt{<person>President Biden</person> spoke in <location>Washington</location> during the <event>campaign</event>.}

\vspace{2mm}
\textbf{OUTPUT FORMAT}

Return only the annotated text, without explanations or additional commentary.

\end{tcolorbox}

\caption{Prompt template used during the entity annotation process.\vspace{-5mm}}
\label{fig:prompt_ner}
\end{figure}

\paragraph{Human Annotation Protocol}
We randomly sampled 105 turns stratified across the 44 debates of \dataset, so that every debate contributes to the sample proportionally to its number of turns. 

Three domain experts, familiar with political discourse and argument mining, independently annotated the sample working only from the raw transcripts, never accessing the LLM, nor each other's annotations.
Annotators followed the same entity inventory $\mathcal{T}$ and the same tagging conventions given to the model (cf. Figure~\ref{fig:prompt_ner}). 

\paragraph{Agreement Values}
Table~\ref{tab:agreement} reports the inter-annotator agreement over the debate-stratified sample of 105 turns, for increasingly heterogeneous annotator pools by combining LLM-based and human-based annotations. Agreement is computed as Fleiss' $\kappa$ over token-level BIO label sequences.

\paragraph{Effect of additional LLM annotators}
We replied the same protocol by considering a second annotator model family, namely, GPT-5.5. As shown in Table~\ref{tab:agreement}, the introduction of these new annotations does not excessively move the agreement values, strengthening our annotation methodology.

\begin{table}[t!]
\centering
\small
\setlength{\tabcolsep}{5pt}
\begin{tabular}{lc}
\toprule
\textbf{Annotator pool} & \textbf{Fleiss' $\kappa$} \\
\midrule
3 human experts & 0.9190 \\
Gemini consensus + 3 humans & 0.9000 \\
3 Gemini runs + 3 humans & 0.8842 \\
3 Gemini + 3 humans + 3 GPT-5.5 & 0.8623 \\
\bottomrule
\end{tabular}
\caption{Token-level BIO inter-annotator agreement over the debate-stratified sample of 105 turns, for increasingly heterogeneous annotator pools.}
\label{tab:agreement}
\end{table}

\paragraph{Recurring Ambiguity Classes}
Inspecting where annotators disagree, we found the corrections to concentrate on a small set of well-known hard cases. First, compound and nested entities, e.g., \texttt{Clinton Foundation}, where \texttt{Clinton} reads as a person and \texttt{Foundation} as an organization, admitting either a split annotation or a nested \texttt{<organization><person>Clinton</person> Foundation</organization>}; the same holds for \texttt{Obama's Medicare}. Second, roles used to denote a specific person, e.g., \texttt{the President}. Third, event-versus-location readings of the same name, e.g., \texttt{Pearl Harbor}. Fourth, the segmentation of titled names, e.g., \texttt{Vice President Biden} as a single \texttt{<person>}, or as \texttt{<role>Vice President</role> <person>Biden</person>}. Notably, these cases were found to be ambiguous for the human experts too, which is why human-human agreement is high but not perfect.

\paragraph{Systematic Tendencies of the LLM Annotators}
By comparing model families and by inspecting where the model diverges from the human annotation, we identified four systematic tendencies of the LLM-assisted pipeline. First, a bias toward finer segmentation: when compared with the GPT-5.5 annotations, Gemini systematically identifies more entities. Second, a very rare generative reformulation, i.e., slight rephrasing rather than verbatim tagging, e.g., \texttt{government} rendered as \texttt{the government}. Third, an occasional splitting of multi-token references into two entities rather than one, e.g., in the case of \texttt{Roe v. Wade}. Fourth, an occasional surface normalization, e.g., capitalizing speaker names given in lowercase, or correcting evident transcription typos, e.g., \texttt{MR. SM1TH} becoming \texttt{MR. SMITH}. 

Crucially, each of these is controlled by our pipeline: reformulation and normalization diverge from the source and are caught and removed by the alignment to the original transcript, while segmentation differences are resolved through the consensus and the human validation pass. As shown in Section~\ref{sec:robustness}, none of them affects the final evaluation.

\section{Additional Results on \dataset}
\label{app:results}

\begin{table*}[t]
\centering
\small
\resizebox{\linewidth}{!}{
\begin{tabular}{lccc|ccccccccc|c}
\toprule
& \multicolumn{3}{c|}{\textbf{Argument Mining}} 
& \multicolumn{9}{c|}{\textbf{DNER}}
& \textbf{Joint}\\
\cmidrule(r){2-4} \cmidrule(l){5-13} \cmidrule(l){14-14}

\textbf{Model} & 
$\rho_{\textsc{claim}}$ & $\rho_{\textsc{premise}}$ & $\rho_{\textsc{AM}}$ & 
$\rho_{\textsc{date}}$ & $\rho_{\textsc{event}}$ & $\rho_{\textsc{law}}$ & $\rho_{\textsc{loc}}$ & $\rho_{\textsc{org}}$ & $\rho_{\textsc{party}}$ & $\rho_{\textsc{person}}$ & $\rho_{\textsc{role}}$ & $\rho_{\textsc{DNER}}$ & $\rho_{\textsc{joint}}$\\
\midrule
Llama 3.1 8B & 0.858 & 0.809 & 0.922 & 0.770 & 0.749 & 0.788 & 0.964 & 0.947 & 0.959 & 0.987 & 0.835 & 0.973 & 0.965\\
Mistral 7B &  0.852 & 0.804 & 0.904 & 0.767 & 0.711 & 0.917 & 0.966 & 0.957 & 0.967 & 0.989 & 0.860 & 0.974 & 0.956 \\
Qwen 2.5 7B & 0.846 & 0.777 & 0.884 & 0.729 & 0.753 & 0.800 & 0.972 & 0.948 & 0.973 & 0.986 & 0.894 & 0.976 & 0.954\\
\bottomrule
\end{tabular}
}
\caption{AM and DNER Pearson corr. between ground-truth and predicted tagged spans in the test set. $\rho_{AM}$, resp. $\rho_{DNER}$ columns are aggregated across all AM, resp. DNER, tags. $\rho_{joint}$ corresponds to the full AM+DNER tags.}
\label{tab:correlation-results}
\end{table*}

\begin{table*}[t!]
\centering
\small
\setlength{\tabcolsep}{3pt}
\resizebox{\linewidth}{!}{
\begin{tabular}{llcccccccccccccccccc}
\toprule
\multirow{2}{*}{\textbf{Approach}} 
& \multirow{2}{*}{\textbf{Model}}
& \multicolumn{9}{c}{\textbf{Arg. Min.}} 
& \multicolumn{9}{c}{\textbf{DNER}} \\

\cmidrule(lr){3-11}
\cmidrule(lr){12-20}

&
& \multicolumn{3}{c}{\textbf{B}}
& \multicolumn{3}{c}{\textbf{I}}
& \multicolumn{3}{c}{\textbf{O}}
& \multicolumn{3}{c}{\textbf{B}}
& \multicolumn{3}{c}{\textbf{I}}
& \multicolumn{3}{c}{\textbf{O}}\\

\cmidrule(lr){3-5}
\cmidrule(lr){6-8}
\cmidrule(lr){9-11}
\cmidrule(lr){12-14}
\cmidrule(lr){15-17}
\cmidrule(lr){18-20}

& 
& P & R & $F_1$
& P & R & $F_1$
& P & R & $F_1$
& P & R & $F_1$
& P & R & $F_1$
& P & R & $F_1$ \\

\midrule

\multirow{3}{*}{\textbf{Ours}}
& Llama 3.1 8B 
& 0.665 & 0.673 & 0.669
& \textbf{0.795} & 0.840 & \textbf{0.817}
& 0.764 & 0.706 & 0.734
& 0.900 & \textbf{0.883} & 0.891
& \textbf{0.893} & \textbf{0.853} & \textbf{0.872}
& \textbf{0.989} & \textbf{0.992} & \textbf{0.990} \\

& Mistral 7B
& \textbf{0.686} & \textbf{0.674} & \textbf{0.679}
& 0.792 & \textbf{0.842} & 0.816
& 0.770 & 0.709 & 0.738 
& \textbf{0.908} & 0.876 & \textbf{0.892}
& 0.861 & 0.845 & 0.853
& \textbf{0.989} & 0.991 & \textbf{0.990}\\

& Qwen 2.5 7B 
& 0.648 & 0.623 & 0.636
& 0.777 & 0.797 & 0.787
& 0.711 & 0.691 & 0.701
& 0.895 & 0.874 & 0.884
& 0.878 & 0.827 & 0.852
& 0.988 & 0.991 & \textbf{0.990}\\

\midrule

\citet{RooseBERT}
& RooseBERT
& 0.107 & 0.198 & 0.139
& 0.723 & 0.765 & 0.743
& 0.777 & 0.653 & 0.710
& 0.472 & 0.463 & 0.467
& 0.466 & 0.359 & 0.406
& 0.953 & 0.962 & 0.957 \\

\midrule

\multirow{3}{*}{\citet{caputo-etal-2026-argument}}
& Llama 3.1 8B 
& 0.305 & 0.438 & 0.360
& 0.484 & 0.741 & 0.585
& 0.662 & 0.391 & 0.492 
& -- & -- & -- 
& -- & -- & --
& -- & -- & -- \\

& Mistral 7B
& 0.208 & 0.328 & 0.255
& 0.359 & 0.584 & 0.445
& 0.731 & 0.521 & 0.608 
& -- & -- & -- 
& -- & -- & --
& -- & -- & -- \\

& Qwen 2.5 7B 
& 0.524 & 0.385 & 0.444
& 0.707 & 0.720 & 0.714
& 0.644 & 0.647 & 0.645
& -- & -- & -- 
& -- & -- & --
& -- & -- & -- \\

\midrule

\multirow{3}{*}{Zero-shot}
& Llama 3.1 8B 
& 0.073 & 0.039 & 0.051
& 0.499 & 0.180 & 0.264
& 0.511 & \textbf{0.826} & 0.631 
& 0.179 & 0.292 & 0.222
& 0.246 & 0.356 & 0.291
& 0.947 & 0.903 & 0.925 \\

& Mistral 7B
& 0.037 & 0.047 & 0.041
& 0.460 & 0.760 & 0.573
& 0.772 & 0.454 & 0.572 
& 0.059 & 0.027 & 0.037
& 0.054 & 0.045 & 0.049
& 0.939 & 0.965 & 0.952 \\

& Qwen 2.5 7B 
& 0.130 & 0.095 & 0.110
& 0.613 & 0.642 & 0.627
& 0.585 & 0.570 & 0.577 
& 0.084 & 0.064 & 0.072
& 0.093 & 0.101 & 0.097
& 0.918 & 0.928 & 0.923 \\

\midrule

\multirow{3}{*}{Few-shot}
& Llama 3.1 8B 
& 0.209 & 0.259 & 0.231
& 0.544 & 0.637 & 0.587
& \textbf{0.816} & 0.744 & \textbf{0.778}
& 0.367 & 0.561 & 0.444
& 0.434 & 0.492 & 0.461
& 0.978 & 0.958 & 0.968\\

& Mistral 7B
& 0.226 & 0.183 & 0.202
& 0.595 & 0.495 & 0.540
& 0.468 & 0.574 & 0.516
& 0.589 & 0.394 & 0.472
& 0.451 & 0.323 & 0.377
& 0.943 & 0.972 & 0.958\\

& Qwen 2.5 7B 
& 0.335 & 0.108 & 0.163
& 0.625 & 0.235 & 0.341
& 0.432 & 0.816 & 0.565 
& 0.773 & 0.432 & 0.554
& 0.649 & 0.403 & 0.497
& 0.948 & 0.987 & 0.967\\

\bottomrule
\end{tabular}
}
\caption{Boundary-only BIO AM performance on the test set of \dataset. \citet{caputo-etal-2026-argument} is not reported because it does not produce joint ACS+ACC labels. Bolded values correspond to the best performance.}
\label{tab:untyped-bio-results}
\end{table*}

\begin{table*}[t!]
\centering
\small
\setlength{\tabcolsep}{3pt}
\begin{tabular}{lccccccccccccccc}
\toprule
\multirow{2}{*}{\textbf{Approach}} &
\multicolumn{3}{c}{\textbf{B-Claim}} &
\multicolumn{3}{c}{\textbf{B-Premise}} &
\multicolumn{3}{c}{\textbf{I-Claim}} &
\multicolumn{3}{c}{\textbf{I-Premise}} &
\multicolumn{3}{c}{\textbf{O}} \\

\cmidrule(lr){2-4}\cmidrule(lr){5-7}\cmidrule(lr){8-10}\cmidrule(lr){11-13}\cmidrule(lr){14-16}

& P & R & $F_1$
& P & R & $F_1$
& P & R & $F_1$
& P & R & $F_1$
& P & R & $F_1$ \\
\midrule

Our
& 0.539 & \textbf{0.541} & \textbf{0.540}
& \textbf{0.476} & \textbf{0.456} & \textbf{0.466}
& 0.580 & \textbf{0.636} & 0.607
& 0.560 & \textbf{0.576} & \textbf{0.568}
& \textbf{0.770} & 0.709 & 0.738 \\

AM-only
& \textbf{0.541} & 0.522 & 0.531
& 0.473 & 0.437 & 0.454
& \textbf{0.616} & 0.615 & \textbf{0.616}
& 0.569 & 0.547 & 0.558
& 0.738 & 0.760 & \textbf{0.749} \\

AM$\rightarrow$DNER
& 0.515 & 0.459 & 0.485
& 0.457 & 0.389 & 0.420
& 0.609 & 0.565 & 0.586
& 0.564 & 0.502 & 0.531
& 0.692 & \textbf{0.779} & 0.733\\

DNER$\rightarrow$AM
& 0.514 & 0.464 & 0.488
& 0.434 & 0.360 & 0.394
& 0.600 & 0.569 & 0.584
& \textbf{0.570} & 0.469 & 0.515
& 0.672 & \textbf{0.779} & 0.721 \\

\bottomrule
\end{tabular}

\caption{
Typed BIO token-level ablation study comparing JAET against AM-DNER pipelines and single-task variants on AM under typed and boundary-only settings on the test set. Bolded values indicate the best performance.}
\label{tab:ablation-typed}
\end{table*}

\paragraph{Component- and entity-count correlations}
First, we investigate whether our proposed approach recovers the correct amount of components and entities from each turn, by computing the component- and entity-count correlations as:
\begin{equation}
\rho_{\text{A}} =
\mathrm{corr}\left(
(|A^*(u)|)_{u\in\mathcal{U}},
(|A^\tau(u)|)_{u\in\mathcal{U}}
\right),
\end{equation}
\begin{equation}
\rho_{\text{E}} =
\mathrm{corr}\left(
(|E^*(u)|)_{u\in\mathcal{U}},
(|E^\tau(u)|)_{u\in\mathcal{U}}
\right),
\end{equation}
where $A$ and $E$ denote the argument component spans and debate named entity spans, respectively, and $\mathrm{corr}$ is the Pearson correlation coefficient, and provides quantitative insights into the annotation process. As shown in Table~\ref{tab:correlation-results}, all considered models achieve high correlation overall, with minor variations for certain debate entities.

\paragraph{Detailed BIO evaluations}
Table~\ref{tab:untyped-bio-results} reports the untyped BIO evaluation on AM and DNER, complementing Table~\ref{tab:main-results-details} from the main text. Moreover, Table~\ref{tab:ablation-typed} provides token-level BIO insights into the ablation study we performed on JAET (cf. Table~\ref{tab:ablation-combined} in the main text). Finally, Table~\ref{tab:preserved} compares JAET and competing methods by only considering turns that have been correctly preserved after tagging.

\begin{table}[t!]
\centering
\small
\setlength{\tabcolsep}{2pt}
\begin{tabular}{lcccc}
\toprule
\multirow{2}{*}{\textbf{Preserved turns only}} & \multicolumn{2}{c}{\textbf{Arg. Min.}} & \multicolumn{2}{c}{\textbf{Joint}} \\
\cmidrule(lr){2-3}\cmidrule(lr){4-5}
 & Untyped & Typed & Untyped & Typed \\
\midrule
RooseBERT           & 0.6200 & 0.4727 & 0.4543 & 0.1218 \\
JAET --- Llama 3.1 8B & 0.7754 & 0.5981 & 0.6694 & 0.4948 \\
JAET --- Mistral 7B   & \textbf{0.8121} & \textbf{0.6516} & \textbf{0.7113} & \textbf{0.5653} \\
JAET --- Qwen 2.5 7B  & 0.7900 & 0.6243 & 0.6847 & 0.5282 \\
\bottomrule
\end{tabular}
\caption{Macro $F_1$ recomputed over correctly-preserved turns only (cf. Section~\ref{sec:results}). Bolded values indicate the best performance.}
\label{tab:preserved}
\end{table}

\begin{table}[t!]
\centering
\small
\setlength{\tabcolsep}{4pt}
\begin{tabular}{lccc}
\toprule
\textbf{Macro $F_1$} & \textbf{Turn-level} & \textbf{Debate-level} & $\Delta_{\text{rel}}$ \\
\midrule
AM (untyped)    & 0.7446 & 0.7266 & $-2.4\%$ \\
AM (typed)      & 0.5837 & 0.5697 & $-2.4\%$ \\
Joint (untyped) & 0.6436 & 0.6114 & $-5.0\%$ \\
Joint (typed)   & 0.4636 & 0.4471 & $-3.6\%$ \\
\bottomrule
\end{tabular}
\caption{Effect of a strict debate-level split for \dataset (35 training / 9 held-out debates) on our reference model, w.r.t. the turn-level split of Table~\ref{tab:dataset-stats}.}
\label{tab:robustness}
\end{table}

\begin{table*}[ht!]
\centering
\small
\setlength{\tabcolsep}{5pt}
\begin{tabular}{llccccccc}
\toprule
\multirow{2}{*}{\textbf{Approach}} & \multirow{2}{*}{\textbf{Model}} &
\multicolumn{2}{c}{\textbf{Arg. Min.}} & \multicolumn{2}{c}{\textbf{Joint}} & \multicolumn{2}{c}{\textbf{Sanity}} \\
\cmidrule(lr){3-4}\cmidrule(lr){5-6}\cmidrule(lr){7-8}
 & & Untyped & Typed & Untyped & Typed & $T_{\mathrm{PR}}$ & $T_{\mathrm{WR}}$ \\
\midrule
\multirow{3}{*}{\textbf{Ours}}
& Llama 3.1 8B & 0.8479 & 0.7463 & 0.5965 & 0.3515 & \textbf{0.8625} & \textbf{0.9875} \\
& Mistral 7B   & \textbf{0.8641} & \textbf{0.7550} & \textbf{0.6844} & \textbf{0.4052} & \textbf{0.8625} & \textbf{0.9875} \\
& Qwen 2.5 7B  & 0.7982 & 0.6850 & 0.5058 & 0.2153 & 0.7750 & 0.9625 \\
\midrule
\multirow{3}{*}{Zero-shot}
& Llama 3.1 8B & 0.2342 & 0.1539 & 0.0956 & 0.0140 & 0.0500 & 0.7625 \\
& Mistral 7B   & 0.3521 & 0.2243 & 0.1038 & 0.0176 & 0.0000 & 0.7125 \\
& Qwen 2.5 7B  & 0.3984 & 0.2757 & 0.1602 & 0.0318 & 0.2875 & 0.7875 \\
\midrule
\multirow{3}{*}{Few-shot}
& Llama 3.1 8B & 0.3716 & 0.2919 & 0.2040 & 0.1359 & 0.5125 & 0.9625 \\
& Mistral 7B   & 0.2478 & 0.1959 & 0.1104 & 0.0435 & 0.3625 & 0.9625 \\
& Qwen 2.5 7B  & 0.4238 & 0.3311 & 0.1942 & 0.0947 & 0.5875 & 0.9750 \\
\bottomrule
\end{tabular}
\caption{Token-level BIO results on the test set of Persuasive Essays~\cite{stab2017persuasiveessay}, under both the boundary-only (untyped) and the typed settings. Bolded values correspond to the best performance.}
\label{tab:pe-full}
\end{table*}

\begin{table}[ht!]
\centering
\small
\setlength{\tabcolsep}{4pt}
\begin{tabular}{lcccc}
\toprule
\multirow{2}{*}{\textbf{Approach}} & \multicolumn{2}{c}{\textbf{Arg. Min.}} & \multicolumn{2}{c}{\textbf{Joint}} \\
\cmidrule(lr){2-3}\cmidrule(lr){4-5}
 & Untyped & Typed & Untyped & Typed \\
\midrule
JAET (joint)      & 0.8641 & 0.7550 & \textbf{0.6844} & \textbf{0.4052} \\
AM-only           & \textbf{0.8759} & \textbf{0.7684} & -- & -- \\
AM$\rightarrow$DNER & 0.8148 & 0.7113 & 0.4814 & 0.2264 \\
DNER$\rightarrow$AM & 0.8507 & 0.7511 & 0.5408 & 0.2654 \\
\bottomrule
\end{tabular}
\caption{Ablation study on Persuasive Essays, replicating the setting of Table~\ref{tab:ablation-combined}. ``--'' indicates unavailable outputs from that method. Bolded values indicate the best performance. \vspace{-1mm}}
\label{tab:pe-ablation}
\end{table}

\section{Additional Results on Robustness}
\label{app:robustness}
\paragraph{Generalization to Unseen Debates}
Table~\ref{tab:robustness} reports the debate-level split discussed in Section~\ref{sec:robustness}, in which 35 debates from \dataset are used for training and 9 entirely held-out debates for testing, so that no turn from a test debate is ever seen during training.

\paragraph{Domain Transfer}
Table~\ref{tab:pe-full} reports the complete evaluation on Persuasive Essays~\cite{stab2017persuasiveessay}, as discussed in Section~\ref{sec:robustness}. We produced the entity layer for this corpus using the same annotation pipeline of Appendix~\ref{app:annotation}, yielding the following inventory: \textsc{Date}, \textsc{Event}, \textsc{Facility}, \textsc{Language}, \textsc{Law}, \textsc{Location}, \textsc{Norp}, \textsc{Organization}, and \textsc{Person} and we mapped \textsc{MajorClaims} onto \textsc{Claims} to match the argument inventory $\mathcal{C}$ of Definition~\ref{def:argument-component}. 
It should be noted that we deliberately do not include the AM-only models of~\citet{caputo-etal-2026-argument} in this comparison, as their released models were originally trained on Persuasive Essays. 
Finally, as in the case of \dataset, Table~\ref{tab:pe-ablation} confirms the gain of JAET cannot be achieved by concatenating sequential AM--DNER pipelines.

\paragraph{Effect of the Turn-level Window}
Table~\ref{tab:window} reports the effect of evaluating our reference model on inputs aggregating an increasing number of consecutive turns (cf. Limitations) from \dataset. 

Both tagging quality and structural fidelity degrade monotonically with the window size, i.e., feeding the model more context at once does not help, and processing the whole debate at once would be ineffective.

\begin{table}[t!]
\centering
\small
\setlength{\tabcolsep}{3pt}
\begin{tabular}{lccccc}
\toprule
\textbf{Window size} & \textbf{1} & \textbf{5} & \textbf{10} & \textbf{20} & \textbf{50} \\
\midrule
AM (untyped)    & \textbf{0.7446} & 0.6531 & 0.5234 & 0.3404 & 0.2111 \\
AM (typed)      & \textbf{0.5837} & 0.5043 & 0.3931 & 0.2372 & 0.1296 \\
Joint (untyped) & \textbf{0.6436} & 0.5478 & 0.4292 & 0.2651 & 0.0761 \\
Joint (typed)   & \textbf{0.4636} & 0.3866 & 0.2751 & 0.1701 & 0.0166 \\
\midrule
$T_{\mathrm{PR}}$ & \textbf{0.959} & 0.6946 & 0.3473 & 0.0357 & 0.0294 \\
$T_{\mathrm{WR}}$ & \textbf{0.998} & 0.9790 & 0.9521 & 0.8690 & 0.8529 \\
\bottomrule
\end{tabular}
\caption{Macro $F_1$ and structural fidelity of our reference model when the input aggregates an increasing number of consecutive turns. Best values are bolded.}
\label{tab:window}
\end{table}

\begin{table*}[t!]
\centering
\small
\renewcommand{\arraystretch}{1.24}
\begin{tabular}{@{}l@{\hspace{1.5em}}p{0.82\linewidth}@{}}
\toprule
\multicolumn{2}{@{}l}{\textbf{Example 1} --- each pipeline recovers exactly one layer.} \\
\midrule
Gold & \texttt{<person>TRUMP</person>: <claim><location>Iran</location> is taking over <location>Iraq</location></claim>.} \\
JAET & \texttt{<person>TRUMP</person>: <claim><location>Iran</location> is taking over <location>Iraq</location></claim>.} \\
AM$\to$DNER & \texttt{<person>TRUMP</person>: <location>Iran</location> is taking over <location>Iraq</location>.} \\
DNER$\to$AM & \texttt{<person>TRUMP</person>: <claim>Iran is taking over Iraq</claim>.} \\
\midrule
\multicolumn{2}{@{}l}{\textbf{Example 2} --- the second stage destroys what the first stage got right.} \\
\midrule
Gold & \texttt{<person>BIDEN</person>: Can I clarify this? <claim>That's just simply not true about <person>Barack Obama</person></claim>. <claim>He did not say sit down with <person>Ahmadinejad</person></claim>.} \\
JAET & \textit{(matches gold)} \\
AM$\to$DNER & \texttt{<person>BIDEN</person>: Can I clarify this? <claim>That's just simply not true about</person> <person>Barack Obama</person>. <claim>He did not say sit down with</person> <person>Ahmadinejad</person>.} \\
DNER$\to$AM & \texttt{<person>BIDEN</person>: Can I clarify this? That's just simply not true about <person>Barack Obama</person>. <claim>He did not say sit down with Ahmadinejad</claim>.} \\
\midrule
\multicolumn{2}{@{}l}{\textbf{Example 3} --- both pipelines return entities and no argumentative structure.} \\
\midrule
Gold & \texttt{<person>ROMNEY</person>: <date>2014</date>. <premise>When you come out in <date>2014</date> I presume I'm going to be <role>president</role></premise>. <claim>I'm going to make sure you get a job</claim>. Thanks <person>Jeremy</person>.} \\
JAET & \textit{(matches gold)} \\
AM$\to$DNER & \texttt{<person>ROMNEY</person>: <date>2014</date>. When you come out in <date>2014</date>, I presume I'm going to be <role>president</role>. I'm going to make sure you get a job. Thanks <person>Jeremy</person>.} \\
DNER$\to$AM & \textit{(identical to AM$\to$DNER)} \\
\midrule
\multicolumn{2}{@{}l}{\textbf{Example 4} --- type confusion the joint model avoids.} \\
\midrule
Gold & \texttt{<person>ROMNEY</person>: <claim>Production on government land of oil is down 14 percent</claim>.} \\
JAET & \texttt{<person>ROMNEY</person>: <claim>Production on government land of oil is down 14 percent</claim>.} \\
AM$\to$DNER & \texttt{<person>ROMNEY</person>: <premise>Production on government land of oil is down 14 percent</premise>.} \\
DNER$\to$AM & \texttt{<person>ROMNEY</person>: <premise>Production on government land of oil is down 14 percent</premise>.} \\
\bottomrule
\end{tabular}
\caption{Representative test-set outputs of our reference model and of the two sequential pipelines, illustrating the two failure mechanisms of sequential composition.}
\label{tab:qualitative}
\end{table*}

\begin{figure*}[t!]
    \centering
    \includegraphics[width=\linewidth]{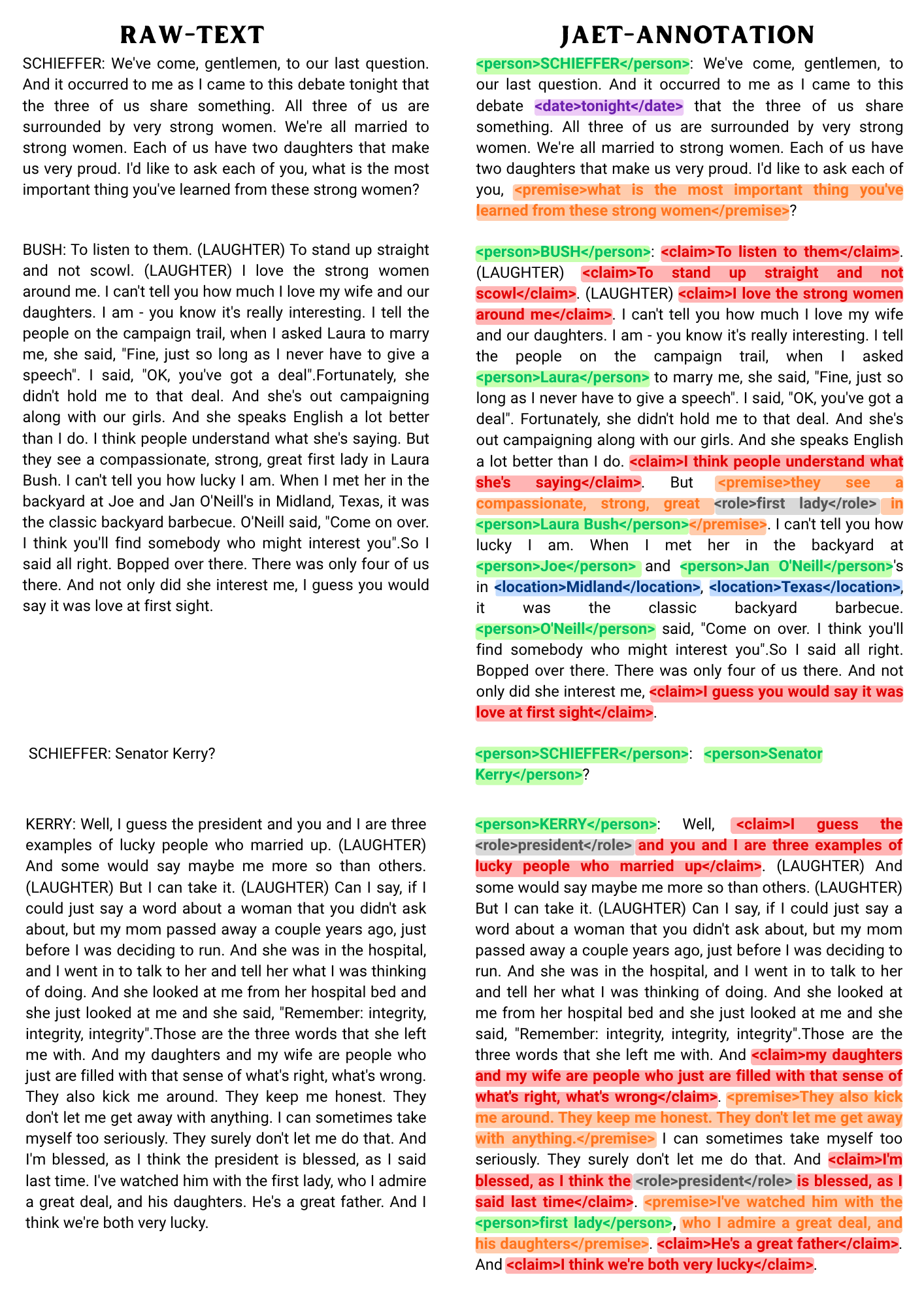}
    \caption{Illustrative example of a debate held on October 13, 2004, with its corresponding JAET annotation.}
    \label{fig:example-1}
\end{figure*}

\section{Qualitative Insights}
\label{app:qualitative}

Table~\ref{tab:ablation-combined} shows that the performance of JAET cannot be recovered by composing two independent modules. In this regard, we isolated two structurally distinct reasons for which sequential pipelines fail:

\noindent $\bullet$ \textit{Each stage is blind to the layer the other predicts:} An entity in subject position signals a predication, and hence a claim; conversely, a claim constrains which entity types are plausible inside it. Neither module can exploit the other's signal, since each is trained in isolation.

\noindent $\bullet$ \textit{The second stage rewrites already-tagged text:} The second stage of a pipeline must insert its own tags into a text that already carries the tags of the first one, a condition it is never supervised on. This produces crossing, ill-formed markup and, more insidiously, the silent deletion of tags the first stage had produced correctly. A single-pass model cannot make this class of error at all.

Table~\ref{tab:qualitative} reports four representative cases. In Example~1, each pipeline recovers precisely the layer its final stage was trained to produce, while the joint model recovers both. Example~2 shows that AM$\to$DNER emits \texttt{</person>} before any \texttt{<person>} opens, producing crossing markup and leaving both claims unclosed; DNER$\to$AM is even more revealing, as its entity stage tags both persons correctly, and its argument stage then keeps \texttt{<person>Barack Obama</person>}, which lies \textit{outside} the claim it failed to produce, and destroys \texttt{<person>Ahmadinejad</person>}, which lies \textit{inside} the claim it did produce. The same pattern appears in Example~1, where the two \texttt{<location>} tags destroyed by the AM stage are exactly the two lying inside the claim it wrapped. In other words, when the argument stage wraps a span, the entity tags nested within it are lost, and the pipeline discards its own correct predictions precisely at the nested argument-entity configurations that motivate this work. In Example~3 both pipelines tag \texttt{<date>}, \texttt{<role>} and \texttt{<person>} correctly, yet emit zero argument components, whereas JAET recovers both the premise and the claim. Finally, Example~4 illustrates the lack of comprehensive signal among the two stages: a statistic advanced as a position is a claim, not a premise, and both pipelines mislabel it, while the joint model does not.

Finally, Figure~\ref{fig:example-1} provides a complete illustrative example of JAET annotation, showing how argument and entity tags are jointly inserted while preserving the original debate transcript.

\end{document}